\documentclass[a4paper]{article}

\usepackage[english]{babel}
\usepackage[utf8x]{inputenc}
\usepackage[T1]{fontenc}
\usepackage{newtxmath,newtxtext}
\usepackage{indentfirst}

\usepackage{mathptmx}
\usepackage[a4paper,top= 1 in,left=1 in,total={6in, 8in}]{geometry}

\usepackage{amsmath}
\usepackage{graphicx}
\usepackage[colorinlistoftodos]{todonotes}
\usepackage[colorlinks=true, allcolors=blue]{hyperref}
\usepackage{marginnote}

\newcommand{\figref}[1]{\figurename~\ref{#1}}
\usepackage{caption} 
\addto\captionsenglish{\renewcommand{\figurename}{FIGURE}}
\addto\captionsenglish{}

\usepackage{fancyhdr}
\title{Beyond Grand Theft Auto V for Training, Testing and Enhancing Deep Learning in Self Driving Cars}
\author{
{Mark Anthony Martinez II} \\ 
{Princeton University} \\
{35 Olden Street, Princeton, NJ 08540} \\
{T: +01 856-701-4511; Email: mam17@cs.princeton.edu}\\
{corresponding author}\\
\\
  {Chawin Sitawarin}\\
  {Princeton University} \\
  {3687 Frist Center, Princeton, NJ 08544} \\
  {T: +01 646-403-0310 ; Email: chawins@princeton.edu}\\
  \\
  {Kevin Finch}\\
  {Princeton University} \\
  {0338 Frist Center, Princeton, NJ 08544} \\
  {T: +01 908-616-4983; Email: kfinch@princeton.edu} \\
  \\
  {Lennart Meincke}\\
  {NordAkademie} \\
  {Köllner Chaussee 11, 25337 Elmshorn, Germany} \\
  {Email: lennart.meincke@nordakademie.de}\\
  \\
  {Alexander Yablonski}\\
  {Princeton University} \\
  {5168 Frist Center, Princeton, NJ 08544} \\
  {T: +01 610-605-7351;Email: aty@princeton.edu}\\
  \\
  {Alain Kornhauser}\\
  {Princeton University} \\
  {229 Sherrerd Hall, Princeton, NJ 08544} \\
  {T: +01 609-258-4657 F: +01 609-258-1563; Email: alaink@princeton.edu}\\
}

\date{}
\begin{document}
\maketitle

\newpage

\section{Abstract}

As an initial assessment, over 480,000 labeled virtual images of normal highway driving were readily generated in Grand Theft Auto V's virtual environment. Using these images, a CNN was trained to detect following distance to cars/objects ahead, lane markings, and driving angle (angular heading relative to lane centerline): all variables necessary for basic autonomous driving. Encouraging results were obtained when tested on over 50,000 labeled virtual images from substantially different GTA-V driving environments.  This initial assessment begins to define both the range and scope of the labeled images needed for training as well as the range and scope of labeled images needed for testing the definition of boundaries and limitations of trained networks.  It is the efficacy and flexibility of a "GTA-V"-like virtual environment that is expected to provide an efficient well-defined foundation for the training and testing of Convolutional Neural Networks for safe driving.  Additionally, described is the Princeton Virtual Environment (PVE) for the training, testing and enhancement of safe driving AI, which is being developed using the video-game engine Unity. PVE is being developed to recreate rare but critical corner cases that can be used in re-training and enhancing machine learning models and understanding the limitations of current self driving models. The Florida Tesla crash is being used as an initial reference. 
\newpage

\section{Introduction}

\indent Self driving cars have the potential to change the paradigm of transportation. According to the U.S. Department of Transportation National Motor Vehicle Crash Causation Survey~\cite{DOT}, 93\% of all vehicle accidents are influenced by human error while driving; eliminating most of the accidents caused by human error would be a giant boon for public safety. Self driving cars also promote ride-sharing which would reduce environmental impact immensely. Additionally, parked cars take up tremendous space in highly populous cities that could be otherwise used for increased living spaces. \cite{wee} 

There have been many advancements in the self driving car space with many companies investing heavily to be the first player in the market. Google’s Waymo has accrued over 3 million miles autonomously. \cite{waymo} Tesla already has a functional autopilot system that allows a car to stay within lanes while it is driving with minimal human intervention. But as much progress has been made there is much work to reach self driving capabilities.
    
There have been several highly public cases where self driving car technology has failed such as the case in Arizona \cite{isaac2017} and most recently publicized in Florida when a Tesla vehicle collided into a truck on the highway. \cite{boudette_2017} Part of the reasons that the systems set in place to drive did not function correctly is because of a lack of data to account for corner cases.
    
The real world offers only one reality (scenario) at a time that is in many cases extremely challenging and costly to create, replicate and iterate.  Virtual reality, while only approximating reality, does provide the opportunity, if properly designed, to readily create scenarios, readily capture the "labeled" data ("affordances") of those scenarios and effectively investigate neighboring variances of those scenarios.  The availability of such environments to test and understand the robustness and effective range of automated driving approaches is thought to be an effective element in the design of those approaches.  While safe behavior in the real world is the final test, the availability of a robust VR environment is thought to be an very effective tool.
    
In this paper, two virtual systems were used to determine whether labeled virtual data can be used successfully for the training, testing and enhancing of self driving car systems are outlined. A popular video game released in 2013, GTA-V, is used to collect labeled data for highways within the game which are then used to train a CNN to detect multiple affordance variables from an unlabeled image of a highway, including an angle between the car and the road, distances to the lane markings and to cars in front. In a orthogonal project, a virtual environment was developed using Unity. It takes map data from Mapzen to create environments that can be used to test difficult to find corner cases. 

\section{Related Work}

Research on self driving cars has existed starting from the 1930s \cite{vanderbilt_2012} and was more seriously considered in the 1980s with the ALVINN project. In that project, a neural network with one hidden layer was used to map input images directly to steering angles. \cite{NIPS1988_95, Pomerleau:1993:NNP:529461} Later on, Aly created a lane marking detector as well as a small testing dataset on roads in urban settings in 2008.~\cite{DBLP:journals/corr/Aly14}

CNNs have been used in driving like scenarios since 2005 when Muller \cite{NIPS2005_2847} proposed a CNN based off road driving robot, DAVE, that mapped images to steering angles. In a more recent publication in 2015, Huval \cite{DBLP:journals/corr/HuvalWTKSPARMCM15} described a CNN system that detects vehicles and lane markings for highway driving showing that CNNs have promise in autonomous driving.

Virtual data has had a successful history in computer vision. Taylor et al. (2007) used a computer game Half-Life to create a system for evaluating tracking in surveillance systems \cite{Taylor}. In a paper released in 2016 \cite{DBLP:journals/corr/ShafaeiLS16} video game data was used to augment real datasets to provide coverage for scenarios that were difficult to find data for in the real world.

In 2016-2017, there were several publications that used GTA-V and other video games for autonomous driving testing and training \cite{filipowicz, DBLP:journals/corr/ChenSKX15}. In Filipowicz and Liu’s paper \cite{filipowicz} a system using GTA was used to detect the distance to stop signs from an in game image.

Since early on in the development of self-driving technology physical test tracks have been used to evaluate self-driving algorithms~\cite{Toyota}. The final benchmark for system's performance will be real world testing, but with rigorous and complex examples of labeled datasets it is possible to evaluate many corner cases. In 2012 Geiger et al. released the KITTI dataset which has become the preeminent dataset in testing self-driving technologies.~\cite{Geiger} The value in the labeled data stems from the complexity and the number of examples it contains. Just as the KITTI dataset provided immensely useful labeled data to train and test on, it is possible to use a virtual environment to create varied and difficult test cases for self-driving technologies. 

\subsection{Interpreting Image Data}

Traditionally, approaches to model a system's reaction to visual input include behavior reflex and mediated perception. The behavior reflex approach uses learning models which internalize the world model. Pomerleau utilized this method to map images directly to the system's actions such as steering angles and speeds \cite{Pomerleau:1993:NNP:529461}. This scheme extensively exploits the capability of deep learning. Nevertheless, it obscures the decision process being made by the network, leaving only a black-box system that handles exceptionally high-risk tasks. This is certainly an undesirable property to manufacturers and users. 

On the other end of the spectrum, mediated perception detects important features using learned models and then builds a world model based on these features. For example, a mediated perception system uses a combination of many subsystems such as vehicle detectors, and pedestrian detectors to locate as many objects as it can in the driving scene. The approach often creates unnecessary complexity by extracting information irrelevant to the driving task. With consideration for the shortcoming of these approaches, Chen et al.~\cite{DBLP:journals/corr/ChenSKX15} proposed a direct perception approach. 

Direct perception creates a model of the world using a few specific indicators extracted directly from the input data. These indicators, called affordances, are often distances that can be used directly by a simple controller to decide on actions for the system to take in real time. The direct perception approach lies in the middle of the spectrum between the other two approaches and takes advantage of both systems by utilizing the network’s capability of accurately retrieving affordances and at the same time keeping the controller explicit and thus, accountable. 

Using the open source driving virtual environment, TORCS, Chen et al. trained a computer to learn a transformation from images into several meaningful affordance indicators needed for highway driving. Their work demonstrated that this approach works well enough to drive a car in a virtual environment, and generalizes to images in the KITTI dataset, and to driving on US 1 near Princeton. It is believed that an autonomous driving system can be designed with the camera playing the role of the human eye in a direct perception approach.

\section{Calculating affordance variables from images}

\begin{figure}[!htbp]
\centering
\includegraphics[width=0.8\textwidth]{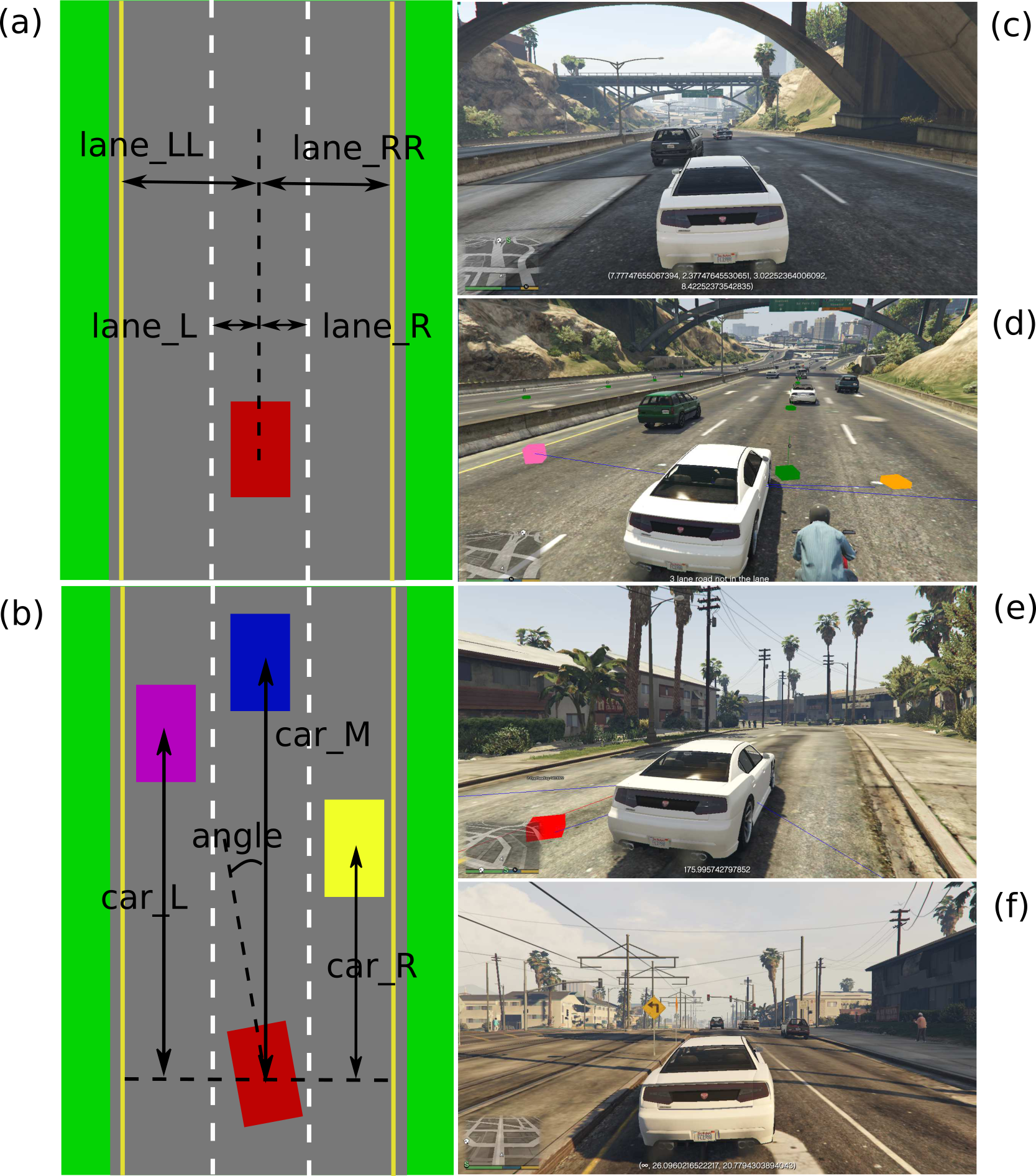}
\caption{(a) A schematic showing 4 affordance variables (lane\_LL, lane\_L, lane\_R, lane\_RR). (b) A schematic showing the other 4 affordance variables (angle, car\_L, car\_M, car\_R). (c) An in-game screenshot showing lane\_LL, lane\_L, lane\_R and lane\_RR. (d)  An in-game screenshot showing the detection of the number of lanes in a road. (e)  An in-game screenshot showing angle. (f)  An in-game screenshot showing car\_L, car\_M and car\_R.}
\label{fig:affordances}
\end{figure}

Using the direct perception model as an inspiration, the following 8 affordance variables for the car were derived from a deep learning model: steering angle (angle), distance to the car in the left lane (car\_L), distance to the car directly in front (car\_M), distance to the car in the right lane (car\_R), distance to the lane marking to the far left (lane\_LL),  distance to the lane marking immediately to the left (lane\_L), distance to the lane marking immediately to the right (lane\_R), and distance to the lane marking to the far right (lane\_RR). In Chen et al., 14 similar variables are chosen; however, some of them appear to be redundant and only complicate the model. \figref{fig:affordances} (a) and (b) are schematics demonstrating the measurement of all 8 affordances.

Some of the affordances (car\_L, car\_M, car\_R, lane\_LL and lane\_RR) might not be applicable in some cases. For example, when there is no car in front, car\_M should not contain any value. Or when the car drives in a two-lane road, either lane\_LL or lane\_RR must be invalid. In this situation, these affordances are set as “inactive.” To indicate the inactive state, the value of these affordances is set to a specific number slightly out of their normal range. This design decision comes into play later on in the training process of the neural network.

\subsection{Learning Model}

The direct perception CNN is based off of the standard single-stream AlexNet architecture.~\cite{alexnet2012} The model is built in Keras using TensorFlow as the backend. Since estimating real values is a regression task instead of an Alexnet's classification based approach, some modifications to the network were made:
\begin{itemize}
\item The shape of the input layer is changed from $224 \times 224 \times 3$ pixels to $210 \times 280 \times 3$ pixels to match the resolution of the screenshots from GTA-V.
\item An extra fully-connected layer is added at the end of the network to serve as the output layer consisting of 8 neurons. Alternatively, instead of adding a layer, one could modify the last layer of AlexNet to have 8 neurons. 
\item The activation function of the output layer must also be changed from softmax to hyperbolic tangent to output a vector of real values in a limited range $[-1, 1]$. 
\end{itemize}
The network is trained with Adam optimizer with an initial learning rate of 0.001, and mean-squared error is chosen as a loss function. The model is trained for 21 epochs under the training set of approximately 350,000 images with a batch size of 32 images. The model quickly converges after the first few epochs.

All input images are mean-subtracted before being fed into the network. Before training and evaluating, all active affordances are rescaled to lie in a range $[-0.9, 0.9]$ to match the output layer of the network, $[-1, 1]$ , and still leave some margin which is necessary to distinguish between active and inactive affordances.

Inactive affordances are rescaled with the same factor. Consequently, they result in a value larger than 1 which is outside of the achievable output range of the network. As a result, the loss will never reach zero regardless of training epochs. Still, this method is desirable for determining whether the affordances are active (i.e. if the output of the network is close to 1, then, with high probability, it is inactive). The inactivity threshold was set to be 0.99, and thus, any output above this threshold is considered inactive.

\subsection{Data Collection Using Grand Theft Auto}

CNNs have shown to be extremely powerful in image classification. The downside that makes CNNs, as well as other supervised machine learning models, often impractical is the lack of appropriate training data. It is not only the dearth of the data, but the quality of the data as well. In order to be effective, machine learning models must be trained on a large number of annotated data that provides enough information for the models to learn from. The greatest obstacle to obtain a usable dataset is often its cost.

There are several major hurdles if one were to use real data for this task. The largest hurdle is the number of data points that would be necessary. In general, 100,000 images would be barely sufficient to train a complicated model. For our purpose, a car would need to drive around recording driving scenes. It is not only expensive to do this manually, but also extremely time consuming. People would have to manually drive a car and collect video data which is limited to geo-location of the car as well working conditions of drivers. 

The other hurdle that one would encounter in using real data, is obtaining the ground truth labels for each image. In our case, the model needs to pick up eight different affordance variables from each image. Even if a car is equipped with the latest in tracking technology it is not possible for a car to pick up the exact distances of cars in front of it as well as the distance to the lanes in all occasions. For instance, lidar is extremely effective but fails when rain is introduced into the scene. The instruments would not be perfectly able to pick up the distances because of normal mechanical error and even if the instruments do pick up a distance to a lane marking or a car there is no assurance that that value will indeed be correct. A human would need to go through the data to annotate the information and look for incorrect labeling which is again extremely time consuming. 

The benefit of GTA-V is that data can be collected without regard to limitations of the real world in terms of time or cost. There is no cost associated with each test run and the game can run indefinitely until the adequate amount of data is collected. Additionally, GTA provides a multitude of different environments that allow the model to train and test on a variety of scenarios that mimic the real world. Of equal importance is that the game will have perfectly annotated data with the ground truth labels. Costly and possibly imprecise sensors are not needed to find the exact measurements of the affordance variables. The labels needed for each image can be found exactly as the game interprets them. 

GTA-V is not normally an academic tool and therefore fan game modification scripts were used to obtain the values for the distances, to set up the simulations, and to run the simulations used for this research. The GTA mod Rage Plugin Hook \cite{rage} was used with collaboration from Lennart Meincke, the creator of the mod, to extract the information from GTA-V to find the affordances. 

In the data collecting process, a simple car model called “Futo” was chosen and a loop which has the character strictly drive on highways to a random position was set up. The car frequently respawns at a random position. Since the data collection takes at least several days of in-game time the data covered all different time of a day. The camera for taking a screenshot is positioned behind a windshield at a similar location to that of a front-facing camera in an autonomous vehicle. 

A screenshot is taken from the window of the game screen, rescaled to $210 \times 280$ pixels and stored as a Bitmap file. One screenshot is taken every 5 iterations of the script’s loop which is approximately 250 ms. Because of how the script works, taking a screenshot every iteration of the loop introduces a higher variation of the interval between each screenshot. For future work a neural network model that incorporates temporal information will be used necessitating the screenshots to be as equally spaced in time as possible.

\subsection{Description of the Affordance Variable Collection}
The affordance variables were calculated by using in-game constructs that were available through the mod. 

The steering angle was easily calculated by comparing the direction of the road with the direction of the car. 
The lanes of the highway were calculated by using GTA-V’s node system that demarcates where roads are. The nodes in the game denote roads, but do not show exactly where lanes are. GTA however indicates whether the road is a one way and if it has two, three, four, or five lanes. Depending on the direction and the number of lanes of the road the node is placed on the road differently. Using this information the distances to each lane marking were calculated using trigonometry.

The distance to the cars in the left, middle, and right lanes were calculated by looking for cars that were in the direction of the next node in front of the camera. This assured that the cars were in front of the camera. To determine whether the cars were in either the left, middle, or right lanes the same calculations were used for determining the lanes markings and made sure that the cars were enclosed within those same lane markings.

\subsection{Diversity of Scenarios}

GTA-V offers a variety of different scenes that can be used for testing. The city of Los Santos, the in game environment, is mostly an urban environment but also includes suburban and rural environments. 

\begin{figure}[!htbp]
\centering
\includegraphics[width=0.8\textwidth]{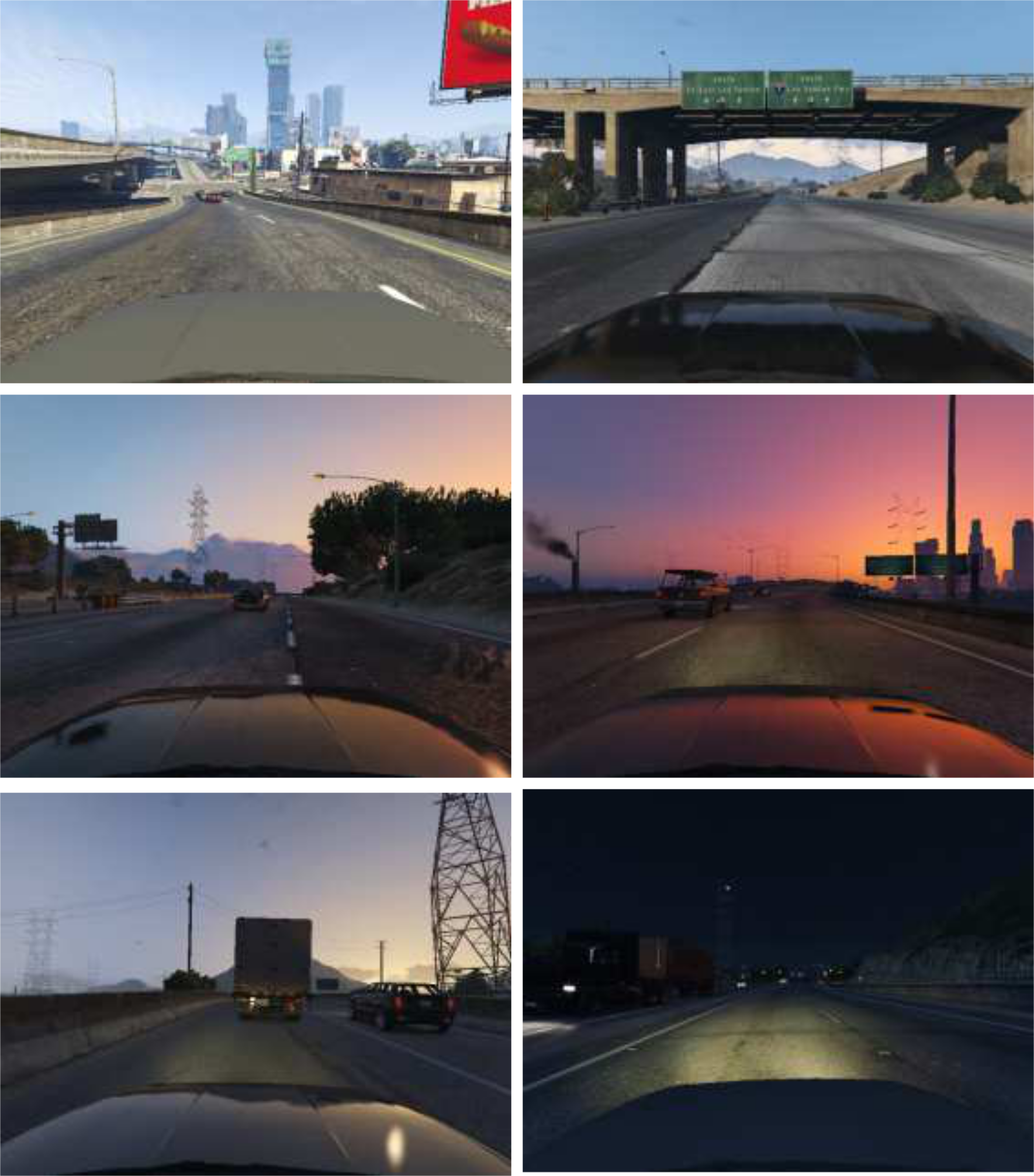}
\caption{Shows different times of the day and different geographic sceneries at which the screenshots are taken}
\label{fig:time_location}
\end{figure}

Additionally GTA-V has a slowly changing time of day that runs 30 times faster than normal time. While collecting data the time of day was not specified and as such collected an evenly distributed amount of data from each in game time.
	Another powerful asset that GTA-V provides is many realistic weather scenarios. At this time different weather patterns were not included in the data and only used what the game describes as “Extra Sunlight”. 

\section{GTA Results}

After the driving simulation in GTA-V had been running for 40 hours, over 500,000 screenshots were taken. After the data cleaning process to remove erroneous screenshots, approximately 483,000 images were left. 350,000 images were separated for training the CNN. The rest were kept for testing and validating purposes. 

The CNN was then trained on a desktop computer running Ubuntu 14.04 equipped with an Nvidia Tesla K40. The training process ran for 21 epochs or approximately 36 hours before manually stopped after showing some sign of overfitting. The training went slightly more slowly than expected because of the bottleneck in reading images stored on a hard disk. The speedup could be achieved by storing the training images in a different database format that requires a shorter access time. 

The final training loss per sample per variable is 0.151 whereas the final validation loss per sample per variable is 0.314. The loss is defined as mean squared error of 8 affordance variables combined. As mentioned earlier, the loss is expectedly high because the values of inactive afforances were set to be above the output range of the CNN. To display the loss in more detail, TABLE \ref{tab:result} shows the mean squared error or the loss of the validation set at various training epochs on each affordance variable separately. The losses shown in TABLE \ref{tab:result} also more accurately represent the true error of the system since they exclude errors that incur from the inactive affordance variables. From TABLE \ref{tab:result}, it is important to note that the losses of all variables decrease as the training progresses. However, the losses of some variables, namely lane\_L, lane\_M and lane\_R, increase after a certain epoch of training, potentially due to overfitting. Regardless, the CNN trained at epoch 21 is chosen as the final model for its overall accuracy.

\begin{table}[!htbp]
\centering
\caption{Mean squared error of the validation set on each affordance variable at different epochs of training}
 \begin{tabular}{|c | c c c c c c c c|} 
 \hline
 Epoch & angle & car\_L & car\_M & car\_R & lane\_LL & lane\_L & lane\_R & lane\_RR \\ [0.5ex] 
 \hline\hline
 6 & 0.008 & 0.441 & 0.618 & 0.138 & 0.114 & 0.030 & 0.012 & 0.064 \\ 
 \hline
 11 & 0.010 & 0.323 & 0.518 & 0.115 & 0.043 & 0.023 & 0.009 & 0.020 \\
 \hline
 16 & 0.009 & 0.322 & 0.491 & 0.109 & 0.038 & 0.024 & 0.014 & 0.018 \\
 \hline
 21 & 0.007 & 0.309 & 0.489 & 0.108 & 0.036 & 0.025 & 0.016 & 0.015 \\
 \hline
\end{tabular}
\label{tab:result}
\end{table}

\begin{figure}[!htbp]
\centering
\includegraphics[width=0.8\textwidth]{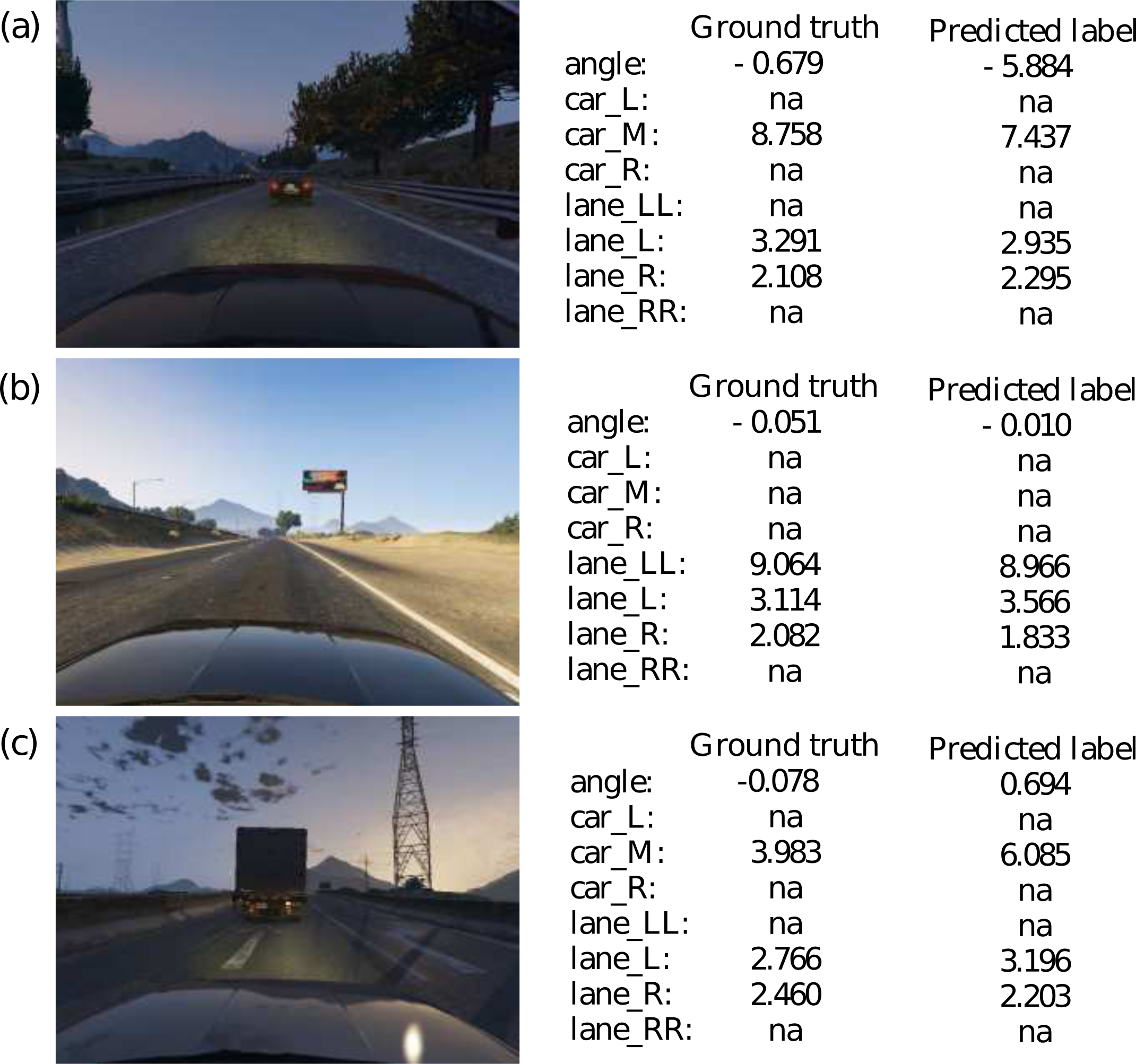}
\caption{Sample images from the test set along with their respective ground truth and predicted labels (na = car/lane does not exist). Please note that angle is in degree and the rest of the variables are in meters.}
\label{fig:samples}
\end{figure}

\figref{fig:samples} also shows some images from the test set, their corresponding ground truth and the predicted label the final model outputs. Generally, as TABLE \ref{tab:result} suggests, the lane distance estimation is more precise than that of the car distance. Nevertheless, when the surrounding cars are closer, the error is arguably small and should not negatively affect the decisions made by the controller. Based on our observation, the large error of car\_M often arises from cars on the left and the right lane at a further distance which are difficult to distinguish. 

\section{Discussion}

We see from the results that the lowest mean squared errors came from the angle of the car to the road and the detection of the lane markings. We theorize that the car distance variables were less successful because the range of possible values and the objects that count as a vehicle is far more varied than lane markings. Lane markings are all in a very similar position in an image and have very little image variation whereas a vehicle could be anything from a motorbike with a pedestrian on it to a freight truck.

As the network trained, the error for car distance continued to decrease with more training while the lane marking estimation started to overfit. A possible solution would be to have two separate neural networks running, one to estimate the car distances and one to estimate the lane marking distance along with the road angle. The disadvantage of this solution is that it requires extra computational power. Another solution might be to use a weighted loss. More precisely, a larger weight can be put on losses incurred from the car distances and a smaller weight on those from the lane marking distances. Doing so could help the car and the lane marking distance estimations converge at a more similar rate.

The error of the car distance estimation is observed to be larger with the cars further away and the cars driving on the left and the right lane. In addition, the error of night drives, in general, tends to be larger than that of drives happen in the day. Certainly, through a longer training, a more extensive training set and a deeper architecture, the network's performance can be significantly improved and thus, the error would be minimized. Nevertheless, it is believed that such properties of the CNN pertains to an inevitable downfall of a system that solely relies on visual inputs. For humans and machines alike, a poor weather and lighting condition as well as a far distance could impair a distance estimation.

Regardless, distances of cars further away and in a different lane affect the controller's decisions in a lower degree compared to those of cars that are closer and in the same lane. Generally, a driving controller maintains the distance to the car directly in front by adjusting the speed accordingly.~\cite{DBLP:journals/corr/ChenSKX15} When a lane change is required, only cars that are close in distance affect the decisions made by the controller. Consequently, the errors in distances of vehicles further, though indeed undesirable, are much more forgivable.

\section{Beyond GTA-V}

\subsection{Limitations of GTA-V}

Though GTA-V provides a great environment for collecting data as mentioned previously, there are drawbacks to the environment as well. First, it was developed and published by Rockstar Games and was not intended to be used as an academic tool. Because of this, there may be legal implications when using GTA-V as a platform for training and testing deep neural networks to drive autonomous vehicles. One such area of concern is the area of liability. 

Additionally, since GTA-V was not developed as a research tool, it is difficult to collect data from the game. Game modification scripts such as Rage Hook are required to even begin to develop ways to collect data from the game, and even then a deep understanding of the game’s APIs and structure is required to develop methods to collect the required affordance indicators. While it is possible to collect the data from video games, it is more beneficial to have a simulator solely dedicated to the purpose of training and testing autonomous vehicles.

This leads to another drawback of GTA-V, its extensibility. While scripts can be written to access certain variables in the game, it does not natively support any add-ons and is difficult to add custom 3D models to the game. Though the environment of GTA-V is large and diverse, there is still only a limited amount of different scenes that can be produced. For example, most areas of the game are an urban setting, and thus, there is a lack of suburban or rural scenes, leading to a relatively limited amount of data being available for those environments.

The other aspect that GTA-V and other simulators lack to some degree is realism. Apart from the quality of the graphics, smaller realistic details should also receive attention, such as motion blur. A car driving at a high speed, especially at night, can introduce a motion blur in an input image while an image captured from GTA-V does not. A motion blur feature is implemented in more recent games and should not be ignored in any realistic driving simulation. In addition, a video taken from a real camera during the night is of lower quality than that of an in-game screen due to sensor artifacts and noises. The varying quality of real inputs can have a strong impact on a system trained only on a virtual environment.

\begin{figure}[!htbp]
\centering
\includegraphics[width=0.8\textwidth]{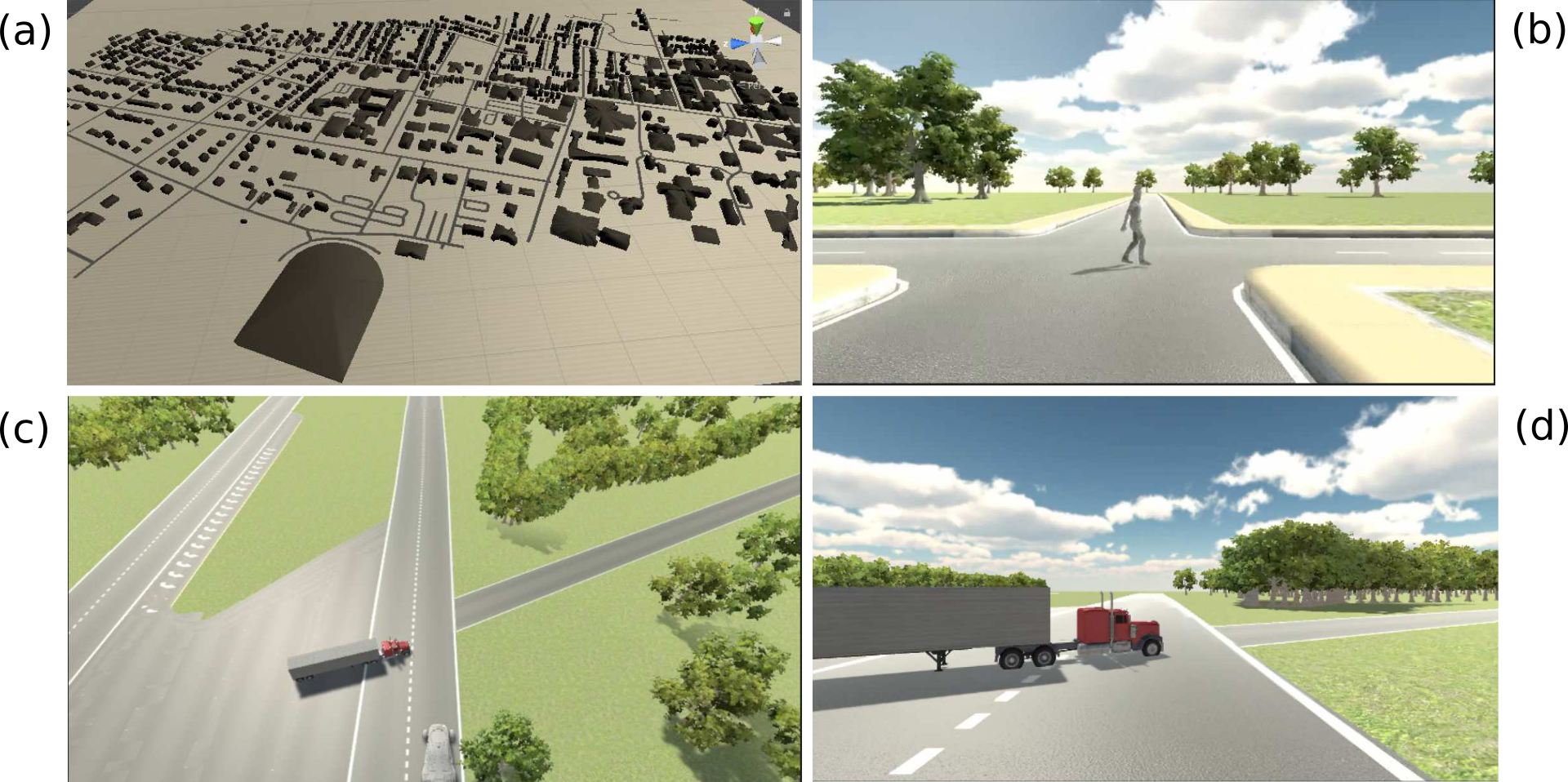}
\caption{(a) An overhead view of a 3D reconstruction of the Princeton area. (b) A car-view of a dangerous driving situation involving a pedestrian crossing the road. (c) An overhead view of a car-truck collision. (d) View of a car-truck collision approaching.}
\label{fig:simulation}
\end{figure}

\subsection{GTA Inspired Research Model}

An ideal virtual environment, then, would be one in which the user has complete control over everything including time of day, weather, road conditions, car models, environments, etc. The ideal system should be easily extensible, allowing the user to import their own 3D models, textures, etc., and allow the user to write scripts to control various aspects of the scenes and the “actors” within it. The user should be able to easily create novel scenes with novel actors, such as the Tesla crash in Florida. Such a system would allow the user to collect substantial data from a limitless set of environments and conditions.

A proof-of-concept virtual environment such as the one just described was developed using the Unity3D game engine \cite{unity}. First, a scene was developed that uses the open data source Mapzen \cite{mapzen} which downloads road and building data for a given latitude/longitude bounding box and renders both the roads and basic 3D models of the buildings that make up that particular bounding box. The data consists of coordinate sets for “nodes” which define roads and coordinate arrays for the base polygons of buildings. The real world coordinates given in the data were converted into suitable game-world coordinates, and the roads were placed into the scene using a script within Unity. In order to create 3D models of buildings, a 3D mesh based on the base polygon of the building and a random height between 5 and 15 meters was created. Figure 4a shows a screenshot of this process, reconstructing the Princeton area. This specific process is useful because it allows the user to create 3D environments based on any location on earth, leading to a huge variety of driving environments which is currently not feasible using a game such as GTA-V. Though there is presently not a large variety of textures for roads and buildings in the current version, the Unity game engine makes it incredibly easy for a user to create their own textures. It is as simple as finding an image online or creating an image with photo editing software and dragging it into Unity.

\subsection{Example of Corner Case: Pedestrian Crossing}

One of the main benefits of using a virtual environment for the collection of data is that no real-world assets or people are harmed in case of a dangerous situation. It is vital for deep neural networks to have data about difficult and rare driving situations such as car accidents, pedestrian interference, etc., but sometimes it is not feasible to recreate those situations in real life. Virtual environments are a perfect place to recreate such edge cases. A scene was created in Unity involving a pedestrian crossing the road in front of the driver’s vehicle, as can be seen in \figref{fig:simulation} (b). This basic scene demonstrates the usefulness of virtual environments. While it would be extremely dangerous to have pedestrians cross in front of cars traveling at various speeds down a road, such an event can be recreated virtually with almost zero cost.

\subsection{Example of Corner Case: Tesla Crash May 7, 2016}

An important aspect of deep neural networks for autonomous vehicles is the ability to retrain and enhance the network with additional data. That data could be specific edge cases that are hard to replicate in GTA-V since it is not a completely open environment. Such an edge case Tesla crash on May 7, 2016. It is vital to be able to gather data from such rare cases in order to enhance the neural networks that our autonomous vehicles will be using. Presently, GTA-V does not have the capability to do such a thing.

Finally, a scene was created to roughly replicate the crash In the scene, shown in \figref{fig:simulation} (c) and (d), you can see the truck turning from Route 27 onto NE 140th Court, and the car approaching that truck. Simulating such a situation in the Unity game engine is useful for many reasons. First, there is a belief that the color of the truck’s trailer led to an issue with the classifier within the Tesla. The simulation easily allows the user to change the color of the truck by adding new textures to Unity, which allows the user to test the effect of various truck colors.

There are several unknown particulars surrounding the Tesla crash, which include the full range of speeds that the truck could have been traveling prior to the crash. For each of those speeds, what could have been seen by both the Tesla driver and the truck driver? Do these views suggest situations where the Tesla driver had no opportunity to stop or avoid the crash? Was the view of the truck driver obstructed (say by the right windshield pillar)? These questions can be answered by recreating the scene in a virtual environment since it allows the user to directly control vehicle speeds, driver viewpoints, etc. Additionally, are there currently existing CNNs or other machine learning models that can implement collision avoidance? If not, can simulations be used to create/test such a model? Recreations of corner cases such as a pedestrian crossing in front of the vehicle and the Tesla crash in Florida represent the usefulness of virtual environments as ones from which data can be collected for use in training deep neural networks, as well as environments in which further study of driving scenarios can be undertaken.

\section{Conclusion}

We examined GTA-V as a model to test, train and enhance deep learning in self driving car research. We found that GTA-V allows researchers to create, train, and test on photo-realistic data to accurately estimate from the driver’s perspective the distance to lane markings, distance to cars, and angle of driving. The best results of the CNN came from estimating the driving angle and the lane markings. The car distance estimation could be improved with more training time, but was terminated before it plateaued because the lane marking detection had started to overfit.

Additionally, we used GTA-V as a model in creating a more robust training environment that would allow researchers to easily create data for testing and training models on. Using our fabricated environment we were able to recreate critical training scenarios, such as the 2016 Tesla crash, that GTA-V would not be able to handle.
    
We showed that virtual data can be a powerful way to create new data invaluable for training and testing neural networks for autonomous driving. The most important way to further validate this work and to make it truly useful is to combine it with real data to understand what the current limitations are with mixing real and virtual data as well as augmenting data sets that need specific corner cases.

\section{Future Work}

We aim to create a more effective deep-learning based autonomous driving system and to develop a virtual environment as a platform for training and testing such systems. The following are prospective directions we wish to take in the near future.

First, we will develop a driving controller based on the direct perception approach which will be tested in both virtual and real-world settings. The model's transferability to real world settings will also be closely investigated. A mixture of real and virtual data in the training set could significantly improve the transferability. 

Second, we speculate that time-dependent deep learning models such as 3D CNNs or recurrent neural networks (RNN), specifically long short-term memory (LSTM) networks, will perform better and are more robust to unseen events than traditional CNNs. Several existing works have shown the models' modest improvement over CNNs in both video classification~\cite{karpathy2014} and in autonomous driving systems.~\cite{Xu2016} However, we believe that, with a large virtual dataset available to us, it is possible to achieve an even more accurate model.

Developing a more realistic simulator is our main long-term goal. As mentioned earlier, while GTA-V offers reasonably realistic graphics and diverse driving scenes, it is not a perfect simulator. An ideal simulator should take into account image noise, motion blur as well as other artifacts from a video input. It should also provide a platform to harvest data with a wide variety of driving scenarios/conditions, corner cases, accidents, etc.

\bibliographystyle{plain}
\bibliography{references.bib}

\begin{thebibliography}{10}

\bibitem{mapzen}
Mapzen. https://mapzen.com/.

\bibitem{rage}
Rage plugin hook. https://ragepluginhook.net/.

\bibitem{unity}
Unity game engine. https://unity3d.com/.

\bibitem{waymo}
Waymo. https://waymo.com/.

\bibitem{DOT}
National motor vehicle crash causation survey, Jul 2008.

\bibitem{DBLP:journals/corr/Aly14}
Mohamed Aly.
\newblock Real time detection of lane markers in urban streets.
\newblock {\em CoRR}, abs/1411.7113, 2014.

\bibitem{boudette_2017}
Neal~E. Boudette.
\newblock Tesla's self-driving system cleared in deadly crash, Jan 2017.

\bibitem{DBLP:journals/corr/ChenSKX15}
Chenyi Chen, Ari Seff, Alain~L. Kornhauser, and Jianxiong Xiao.
\newblock Deepdriving: Learning affordance for direct perception in autonomous
  driving.
\newblock {\em CoRR}, abs/1505.00256, 2015.

\bibitem{filipowicz}
Artur Filipowicz, Jeremiah Liu, and Alain Kornhauser.
\newblock Learning to recognize distance to stop signs using the virtual world
  of grand theft auto 5.

\bibitem{Geiger}
A.~Geiger, P.~Lenz, and R.~Urtasun.
\newblock Are we ready for autonomous driving? the kitti vision benchmark
  suite.
\newblock In {\em 2012 IEEE Conference on Computer Vision and Pattern
  Recognition}, pages 3354--3361, June 2012.

\bibitem{DBLP:journals/corr/HuvalWTKSPARMCM15}
Brody Huval, Tao Wang, Sameep Tandon, Jeff Kiske, Will Song, Joel
  Pazhayampallil, Mykhaylo Andriluka, Pranav Rajpurkar, Toki Migimatsu, Royce
  Cheng{-}Yue, Fernando Mujica, Adam Coates, and Andrew~Y. Ng.
\newblock An empirical evaluation of deep learning on highway driving.
\newblock {\em CoRR}, abs/1504.01716, 2015.

\bibitem{isaac2017}
Mike Isaac.
\newblock Uber suspends tests of self-driving vehicles after arizona crash, Mar
  2017.

\bibitem{karpathy2014}
A.~Karpathy, G.~Toderici, S.~Shetty, T.~Leung, R.~Sukthankar, and L.~Fei-Fei.
\newblock Large-scale video classification with convolutional neural networks.
\newblock In {\em 2014 IEEE Conference on Computer Vision and Pattern
  Recognition}, pages 1725--1732, June 2014.

\bibitem{alexnet2012}
Alex Krizhevsky, Ilya Sutskever, and Geoffrey~E Hinton.
\newblock Imagenet classification with deep convolutional neural networks.
\newblock In F.~Pereira, C.~J.~C. Burges, L.~Bottou, and K.~Q. Weinberger,
  editors, {\em Advances in Neural Information Processing Systems 25}, pages
  1097--1105. Curran Associates, Inc., 2012.

\bibitem{NIPS2005_2847}
Urs Muller, Jan Ben, Eric Cosatto, Beat Flepp, and Yann~L. Cun.
\newblock Off-road obstacle avoidance through end-to-end learning.
\newblock In Y.~Weiss, P.~B. Sch\"{o}lkopf, and J.~C. Platt, editors, {\em
  Advances in Neural Information Processing Systems 18}, pages 739--746. MIT
  Press, 2006.

\bibitem{NIPS1988_95}
Dean~A. Pomerleau.
\newblock Alvinn: An autonomous land vehicle in a neural network.
\newblock In D.~S. Touretzky, editor, {\em Advances in Neural Information
  Processing Systems 1}, pages 305--313. Morgan-Kaufmann, 1989.

\bibitem{Pomerleau:1993:NNP:529461}
Dean~A. Pomerleau.
\newblock {\em Neural Network Perception for Mobile Robot Guidance}.
\newblock Kluwer Academic Publishers, Norwell, MA, USA, 1993.

\bibitem{Taylor}
Geoffrey R.~Taylor, Andrew J.~Chosak, and Paul C.~Brewer.
\newblock Ovvv: Using virtual worlds to design and evaluate surveillance
  systems.
\newblock pages 1 -- 8, 07 2007.

\bibitem{Toyota}
R.~Schmidt, H.~Weisser, P.~Schulenberg, and H.~Goellinger.
\newblock Autonomous driving on vehicle test tracks: overview, implementation
  and results.
\newblock In {\em Proceedings of the IEEE Intelligent Vehicles Symposium 2000
  (Cat. No.00TH8511)}, pages 152--155, 2000.

\bibitem{DBLP:journals/corr/ShafaeiLS16}
Alireza Shafaei, James~J. Little, and Mark Schmidt.
\newblock Play and learn: Using video games to train computer vision models.
\newblock {\em CoRR}, abs/1608.01745, 2016.

\bibitem{vanderbilt_2012}
Tom Vanderbilt.
\newblock Autonomous cars through the ages, Feb 2012.

\bibitem{wee}
Michele~Bertoncello Wee and Dominik.
\newblock Ten ways autonomous driving could redefine the automotive world.

\bibitem{Xu2016}
Huazhe Xu, Yang Gao, Fisher Yu, and Trevor Darrell.
\newblock End-to-end learning of driving models from large-scale video
  datasets.
\newblock {\em CoRR}, abs/1612.01079, 2016.

\end{thebibliography}

\end{document}